%% file: Template.tex
\documentclass{article}
\usepackage{spconf,amsmath,graphicx,hyperref}
\usepackage{multirow}
\usepackage{multicol}
\usepackage{booktabs}
\usepackage{amssymb}
\title{Speech LLMs That Breathe: Modeling Non-Verbal Vocalizations in Dialogue}
\name{%
\shortstack{%
Heyang Liu$^{1,2}$,
Lixin Zhang$^{1}$,
Jiayi Huang$^{1}$,
Ziyang Cheng$^{1,2}$,
Yuhao Wang$^{1}$,\\
Ronghua Wu$^{2}$,
Qunshan Gu$^{2}$,
Yanfeng Wang$^{1}$,
Yu Wang$^{1\dagger}$%
}%
\thanks{$^\dagger$Corresponding author.}%
}

\address{%
$^1$Shanghai Jiao Tong University
\quad
$^2$Ant Group
}

\name{%
\shortstack{%
Heyang Liu$^{1,2}$,
Lixin Zhang$^{1}$,
Jiayi Huang$^{1}$,
Ziyang Cheng$^{1,2}$,
Yuhao Wang$^{1}$,\\
Ronghua Wu$^{2}$,
Qunshan Gu$^{2}$,
Yanfeng Wang$^{1}$,
Yu Wang$^{1\dagger}$%
}%
\thanks{$^\dagger$Corresponding author.}%
}

\address{%
$^1$Shanghai Jiao Tong University
\quad
$^2$Ant Group
}
\begin{document}
%\ninept
%
\maketitle
\begin{abstract}
The user experience of speech interaction models has improved, benefiting from the modeling of paralinguistic attributes and advances in speech reconstruction paradigms. However, human communication follows extensive norms regarding non-verbal vocalizations (NVVs) spanning respiratory, affective, and discourse. The omission of these events imposes inherent limitations on the naturalness of current approaches. In this work, we cover 13 major types of NVVs and construct a natural dialogue dataset that conforms to human conversational norms through an automated synthesis pipeline. We enhance the ability of speech interaction models to generate contextually appropriate NVV events through multi-stage fine-tuning and NVV-aware GRPO (Group Relative Policy Optimization). Based on distinct modeling paradigms, we substantially improve the expressiveness of VocalNet and VITA-Qinyu in empathy interactions and role-play dialogues, and further discuss the architectural necessity for achieving high-quality natural speech generation.
\end{abstract}

\input{latex/1st_introduction}
\input{latex/2nd_related_works}

\input{latex/3rd_method}
\input{latex/4th_experiment}

\input{latex/5th_conclusion}

\bibliographystyle{IEEEbib}
\bibliography{strings,refs}

\end{document}

%% file: latex/1st_introduction.tex
\section{Introduction}
\label{sec:intro}

% Human speech interaction conveys rich information beyond textual content, incorporates speaker-independent timbres, paralinguistic emotional cues, and diverse non-verbal vocalizations (NVV). Represented by sighs, laughter, and affirmations, these events convey the speaker's subjective inclinations and are essential for natural dialogue~\cite{hall2009psychosocial, scherer2011assessing}. Speech interaction models receive users' vocal instructions and generate corresponding spoken responses, moving toward increasingly natural and human-like directions~\cite{xu2025qwen2, wang2025vocalnet, xu2026vita}. Although growing research has focused on incorporating emotional characteristics, the integration of NVV events that conform to human conversational norms remains underexplored.

Human speech interaction conveys rich information beyond textual content, encompassing speaker-specific timbres, paralinguistic emotional cues, and diverse non-verbal vocalizations (NVVs). These characteristics complement the linguistic content and play an important role in conveying information that is difficult to capture through text alone. In particular, NVV events such as sighs, laughter, and affirmations can reflect subjective attitudes and conversational intentions, and are essential to expressive human dialogues~\cite{hall2009psychosocial}. Speech interaction models, especially speech large language models (Speech LLMs), receive users' vocal instructions and generate corresponding spoken responses, moving toward increasingly natural and human-like interactions~\cite{fang2025llama, wang2025vocalnet, xu2026vita}. Beyond accurately generating linguistic content, an ideal Speech LLM should be capable of modeling the rich vocal behaviors that naturally accompany human communication. Recent research has increasingly explored the emotional characteristics, enabling more affective speech generation. However, the integration of NVVs appropriate for conversational norms remains underexplored, and current Speech LLMs still fail to naturally produce conversational vocal behaviors.

% , despite their growing capability in speech understanding and generation.

The advances in speech synthesis models have made it increasingly feasible to develop interactive models capable of producing NVV events. Recent approaches support a predefined set of NVV event types~\cite{du2025cosyvoice}, while the latest closed-source models can incorporate appropriate NVV vocalizations while maintaining high-quality speech generation~\cite{zhang2025minimax}. Natural conversation requires the model to determine when an NVV should occur, where it should be placed, and which type of NVV is appropriate for the conversational context. In this work, we investigate this problem and develop Speech LLMs capable of generating NVVs in accordance with human conversational norms. Specifically, we begin by training a text-based predictor capable of generating NVV tokens using Mandarin real-world speech data. To facilitate spoken interactions, we synthesize natural dialogues related to empathy-oriented interactions and role-play dialogues, and leverage an external LLM to annotate NVV events, with the NVV predictor to assess their appropriateness, and Minimax-HD 2.8 for high-quality speech synthesis. Beyond the conventional two-stage paradigm of speech-to-text and speech-to-speech training, we further introduce NVV-aware GRPO to improve the contextual appropriateness of NVV events in generated responses. From two distinct architectures, we develop VocalNet-NV and VITA-NV, and demonstrate that multi-codebook modeling that preserves fine-grained acoustic information is a preferable choice for high-quality speech interactions. Our contributions are summarized as follows:

% Experiments with two different generation paradigms, VocalNet and VITA-Qinyu, demonstrate that multi-codebook modeling that preserves fine-grained acoustic information is a preferable choice for high-quality speech interactions. Our contributions are summarized as follows:

% Specifically, we focus on the modeling of NVV event selection, contextual placement, and natural vocal realization, aiming to enable more natural and expressive spoken interactions.
% Experiments show that the predictor can effectively assess whether the insertion of an NVV event is appropriate in a given conversational context. 

\begin{itemize}
    \item We construct a data generation pipeline for conversational speech with contextually appropriate NVVs.
    \item We propose NVV-aware GRPO, which improves the appropriateness of model responses under substantial distribution discrepancies among NVV events.
    \item We compare two major speech generation paradigms, revealing the key limitations of single-codebook modeling for natural speech generation.
    \item Extensive experiments demonstrate substantial improvements in the naturalness and human-likeness of responses from both VocalNet-NV and VITA-NV.
\end{itemize}

%% file: latex/2nd_related_works.tex
\section{Related Works}
\label{sec:related_works}

Speech interaction models accept spoken instructions from users and generate coherent, aligned textual and speech responses. Cascaded pipelines are considered suboptimal due to information loss, error propagation, and cumulative latency. In contrast, Speech LLMs achieve direct speech interactions. Representative Speech LLMs, including LLaMA-Omni and VocalNet, adopt single-stream speech tokens~\cite{fang2025llama, wang2025vocalnet}. The LLM backbones autoregressively generate semantic speech tokens and then combine with external tone prompts via flow matching to produce speech responses. Conversely, multi-codebook models, represented by Qwen3-Omni and VITA-Qinyu, simultaneously decode all tokens required for speech reconstruction, which encompasses both high-level semantics and fine-grained acoustic details, such as speech features derived from residual vector quantization (RVQ)~\cite{xu2025qwen3, xu2026vita}. Although both approaches enable seamless speech interaction, they generally lack NVV characteristics of human speech, thereby limiting their naturalness and anthropomorphism. JoyAI-Talker incorporates several NVV tokens. However, it has neither explored their placement appropriateness nor investigated the feasibility across model architectures~\cite{bai2026joyai}.

% whether their positions are natural in conversations nor investigated the feasibility across model architectures~\cite{bai2026joyai}.

%% file: latex/3rd_method.tex
\section{Methods}
\label{sec:methods}

% \subsection{NVV Event Modeling}

% We comprehensively consider a wide range of open-source datasets and merge or distinguish categories according to their original class definitions, retaining 13 NVV categories.

% ; most belong to physiological sounds and discourse interjections, while two categories are closely associated with affective expressions.

\begin{figure*}[h!]
    \centering
    \includegraphics[width=1\linewidth]{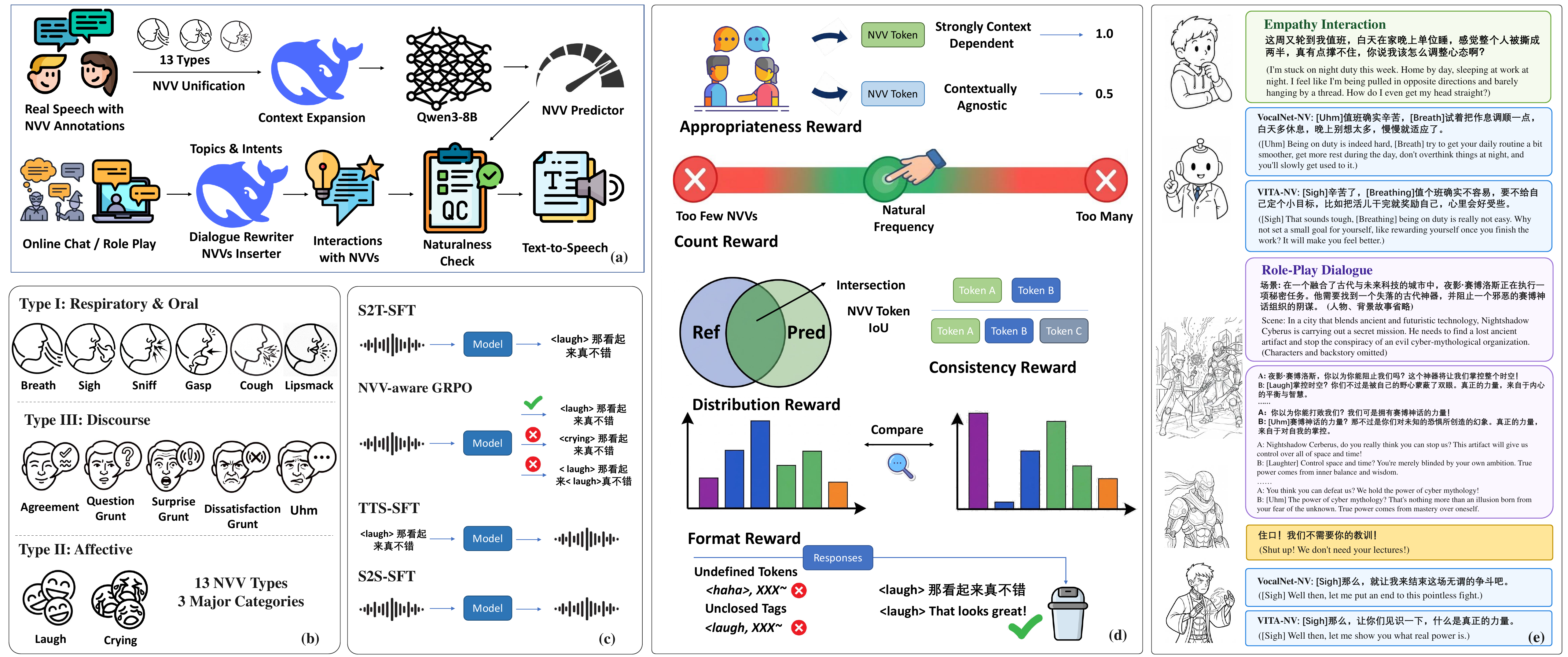}
    \caption{Method overview. (a) Construction pipeline for speech datasets with NVVs; (b) Covered 13 NVV types; (c) NVV-centric multi-stage training; (d) Reward design for NVV-aware GRPO; (e) Interaction examples in both scenarios.}
    \label{fig:main}
\end{figure*}

As shown in Figure~\ref{fig:main} (b), regarding the covered NVV types, we comprehensively consider a wide range of open-source datasets and merge or distinguish categories according to their original class definitions~\cite{wang2024disfluencyspeech, borisov2025nonverbaltts, wu2025smiip, ye2025scalable, liao2026emilia, mai2026mnv}. 

% We ultimately retain 13 NVV categories, most of which belong to physiological sounds and discourse interjections, while laughter and crying are closely associated with affective expressions.

% \begin{table}[h]
% \centering
% \resizebox{0.5\textwidth}{!}{
% \begin{tabular}{cc}
% \toprule
%  \textbf{Category} & \textbf{Supported Types}  \\ \midrule 
%  Respiratory \& Oral &  breath, sigh, sniff, gasp, cough, lipsmack \\ \hline
%  Affective & laughter, crying \\  \hline
%   \multirow{2}{*}{Discourse} & agreement, question\_grunt, surprise\_grunt \\
%   &  dissatisfaction\_grunt, uhm \\
% \bottomrule
% \end{tabular}
% }
% \caption{Covered NVV Types.}
% \label{tab:nvv_types}
% \end{table}

\subsection{NVV Predictor} \label{sec:nvv_predictor}

% Verifying predictable patterns of NVVs in conversations is the prerequisite for achieving natural conversational systems. 

Modeling the contextual regularities of NVVs is essential for natural conversational systems. We utilize extensively collected real-world speech and unify the NVV labels to train Qwen3-8B, enabling it to generate NVV tokens alongside standard text. For isolated single-turn instances, we employ DeepSeek-V4-Pro~\cite{xu2026deepseek} for contextual expansion to provide 2–3 plausible context sentences. The generated NVV predictor is able to distinguish the presence of a natural NVV from random placement. In the synthetic dialogue pipeline, it is employed to assess the plausibility of NVV labels and filter for simulated interactions.

\subsection{Corpus Construction}

% We construct synthesized datasets related to empathy interactions and multi-turn role-play dialogues. 

For empathy interactions, instances are sourced from LCCC~\cite{wang2020large}. DeepSeek-V4-Pro is utilized to construct single-turn interactions with distinct affective inclinations based on topics and user intents. For multi-turn role-play dialogues, our data originates from open-source corpora, retaining solely spoken conversational utterances. DeepSeek-V4-Pro is likewise employed to insert appropriate NVV tokens, and the NVV predictor is subsequently applied to verify their plausibility, yielding natural and pragmatically grounded data.

 % where action descriptions and behavioral instructions on the model response side are stripped, 

% , a large-scale online chat dataset

\subsection{NVV-Centric Multi-stage Training}

\subsubsection{Speech-to-Text SFT}

Given that the vast majority of NVV events are brief and without overlap with regular speech, we extend the model vocabulary with corresponding special tokens and perform supervised fine-tuning (SFT) on synthesized dialogues for text responses. This stage enables the model to generate responses containing NVV events and provides an effective foundation for subsequent NVV-aware reinforcement learning.

\subsubsection{NVV-aware GRPO}

To enhance the model's comprehension of natural dialogues, we conduct NVV-aware GRPO (Group Relative Policy Optimization). The reward function comprises five key aspects:

\vspace{-4mm}
\begin{equation}
\begin{aligned}
    \mathcal{R}_{nvv} ={}&
    \lambda_{\mathrm{appropriate}} \mathcal{R}_{\mathrm{appropriate}}
    + \lambda_{\mathrm{count}} \mathcal{R}_{\mathrm{count}}  \\
    & + \lambda_{\mathrm{consist}} \mathcal{R}_{\mathrm{consist}} + \lambda_{\mathrm{dist}} \mathcal{R}_{\mathrm{dist}}
    + \lambda_{\mathrm{format}} \mathcal{R}_{\mathrm{format}}
\end{aligned}
\end{equation}

\noindent where $\mathcal{R}_{\mathrm{format}}$ indicates format reward:

\vspace{-3mm}
\begin{equation}
\begin{aligned}
    \mathcal{R}_{\mathrm{format}} =
    \begin{cases}
        % -1 & \text{if contains undefined NVV tokens,} \\
           % & \text{unclosed NVV tags, or empty response}, \\
        -1 & \text{undefined NVV tokens, unclosed tags, etc.} \\
         0 & \text{otherwise}.
    \end{cases}
\end{aligned}
\end{equation}

\noindent \textbf{Appropriateness reward:} For appropriateness, we strengthen responses in which the NV token's predicted probability by the predictor surpasses alternative NVV tokens within the given context. Specifically, we downweight contextually agnostic tokens (e.g., breath) to ensure a balanced reward. For context $c$ and token $t$, the score is computed as:

\vspace{-3mm}
\begin{equation}
\begin{aligned}
    \mathcal{S}_{\mathrm{reason(c, t)}} =
    \begin{cases}
        1  &  rank(c,t)=1, t\notin\{breath, ...\}     \\
        0.5 & rank(c,t)=1, t\in\{breath, ...\} \\
        0 & \text{otherwise}.
    \end{cases}
\end{aligned}
\end{equation}

The final reasonableness reward is computed as the average score across all NVV events, with an additional bonus awarded to fully plausible responses containing strongly context-dependent tokens.

\vspace{-4mm}
\begin{equation}
\begin{aligned}
    \mathcal{R}_{\mathrm{reason}}
    &=
    \frac{1}{|\mathcal{C}||\mathcal{T}|}
    \sum_{c\in\mathcal{C}}\sum_{t\in\mathcal{T}}
    \mathcal{S}_{\mathrm{reason}}(c,t) \\
    &\quad
    + b \mathbb I \left[
    \min_{c\in\mathcal{C},\,t\in\mathcal{T}}
    \mathcal{S}_{\mathrm{reason}}(c,t)=1
    \right].
\end{aligned}
\end{equation}

\noindent \textbf{Count reward:} The count reward is designed to penalize the model for aggressively generating excessive relatively safe NVV events, or conversely, outputting too few event tokens to evade potential errors. It is formulated as follows:

\vspace{-2mm}
\begin{equation}
\begin{aligned}
    \mathcal{R}_{\mathrm{count}} = \max\left(0,\, 1 - \frac{\vert{}n_{\text{pred}} - n_{\text{ref}}\vert{}}{\max(n_{\text{ref}},\, 2)}\right)
\end{aligned}
\end{equation}

\noindent \textbf{Consistency reward:} The consistency reward depends on the distance between predicted and reference events and is calculated using class-level IoU (Intersection over Union). It prioritizes accurate predictions compared to reasonable events.

\vspace{-2mm}
\begin{equation}
\begin{aligned}
    \mathcal{R}_{\mathrm{consist}} =  \frac{\text{pred}\cap\text{ref}} {\text{pred}\cup \text{ref}}
\end{aligned}
\end{equation}

\noindent \textbf{Distribution reward:} The distribution reward is to prevent the model from overproducing a particular NVV category and exploiting relatively safe or context-agnostic tokens. Specifically, we measure the discrepancy between the empirical distribution of predicted categories and their natural distribution in the training corpus using the total variation (TV) distance:

\vspace{-4mm}
\begin{equation}
\begin{aligned}
    \mathrm{TV}(P_{\mathrm{pred}}, P_{\mathrm{ref}})
    =
    \frac{1}{2}
    \sum_{c}
    \left|
    P_{\mathrm{pred}}(c) - P_{\mathrm{ref}}(c)
    \right|,
\end{aligned}
\end{equation}

% where $P_{\mathrm{pred}}(c)=n_{\mathrm{pred}}(c)/n_{\mathrm{pred}}$ is the empirical distribution of predicted NVV categories and $P_{\mathrm{ref}}(c)$ is the natural reference distribution estimated from the full SFT dataset. 

\noindent The distribution reward is defined as:

\vspace{-4mm}
\begin{equation}
\begin{aligned}
    \mathcal{R}_{\mathrm{dist}}
    =
    \begin{cases}
        1 - \mathrm{TV}(P_{\mathrm{pred}}, P_{\mathrm{ref}})
        & \text{if } n_{\mathrm{pred}}>0, \\
        0 & \text{otherwise}.
    \end{cases}
\end{aligned}
\end{equation}
\vspace{-2mm}

\begin{table*}[h!]
\vspace{-0.2cm}
\centering
\resizebox{1\textwidth}{!}{
\begin{tabular}{lccccccccccc}
\toprule
 \multirow{3}{*}{\textbf{Base Model}} &  \multirow{3}{*}{\textbf{Phase}}  & \multicolumn{4}{c}{\textbf{Empathy Interaction}}  & \multicolumn{4}{c}{\textbf{Role-play Dialogue}}  \\   \cmidrule(lr){3-6} \cmidrule(lr){7-10} 
 &  & \multicolumn{2}{c}{Appropriateness} & \multicolumn{2}{c}{Consistency} & \multicolumn{2}{c}{Appropriateness} & \multicolumn{2}{c}{Consistency} \\  \cmidrule(lr){3-4} \cmidrule(lr){5-6} \cmidrule(lr){7-8} \cmidrule(lr){9-10} 
 &  & Token (\%) & Response (\%) & Token (\%) & Response (\%) & Token (\%) & Response (\%) & Token (\%) & Response (\%)  \\ \hline
 \multirow{3}{*}{VocalNet} & S2T-SFT & 92.5 & 85.2 & \textbf{55.9} & \textbf{22.4} & 97.1 & 96.8 & 38.5 & 25.3  \\
  & NVV-GRPO & 96.8 & 93.5 & 55.2 & 21.3 & \textbf{97.5}  & 96.9 & 34.1 & 25.8 \\ 
  & S2S-SFT & \textbf{98.0} & \textbf{96.0} & 55.5 & 20.2 & 97.2 & \textbf{97.0} & \textbf{43.8} & \textbf{32.2} \\ \hline
 \multirow{3}{*}{VITA-Qinyu} & S2T-SFT & 94.6 & 89.6 & 51.3 & 17.2 & 95.2 & 94.5 & 40.9 & 30.0 \\
  & NVV-GRPO & \textbf{95.2} & \textbf{91.0} & 52.5 & 17.0 & \textbf{96.2} & \textbf{96.0} & \textbf{44.1} & \textbf{33.0} \\ 
  & S2S-SFT & 92.2 & 84.6 & \textbf{52.9} & \textbf{18.1} & 95.6 & 95.0 & 41.5 & 27.0 \\
\bottomrule
\end{tabular}
}
\caption{NVV appropriateness and reference consistency across training stages.}
\vspace{-2mm}
\label{tab: nvv_position}
\end{table*}

\noindent Using the full training dataset to construct the reference distribution prevents the reward from inheriting category biases introduced by over-sampling in the GRPO training set. Consequently, the reward penalizes excessive concentration on any single NVV category while encouraging the model to preserve a natural category distribution.

\subsubsection{Speech-to-Speech SFT}

This stage trains the model to incorporate fluent NVV events into spoken responses. Given the limited speaker diversity and scale of the synthesized data, we leverage open-source data for text-to-speech training to capture general representations of NVVs~\cite{wang2024disfluencyspeech, borisov2025nonverbaltts, wu2025smiip, ye2025scalable, liao2026emilia, mai2026mnv}. Subsequently, the synthesized dialogue data is used for further training.

%% file: latex/4th_experiment.tex
\section{Experiments}
\label{sec:exp}

\subsection{Non-verbal Token Prediction}

We utilize Emilia-NV~\cite{liao2026emilia} and nonverbalspeech38k~\cite{ye2025scalable} to train Qwen3-8B to generate NVV tokens alongside standard text, as explained in Section~\ref{sec:nvv_predictor}. 200K instances are used for training. The evaluation set comprises 6,034 recordings spanning over 317K token positions with around 7K real NVV occurrences. For each position, the fine-tuned model is employed to predict the presence of an NVV, achieving an AUC-ROC of 0.7635. Furthermore, at real NVV positions, the median predicted probability is 75.0 times higher than non-NVV positions, demonstrating the effective capture of the vocalization distributions in natural conversations. Across the considered NVV categories, the ground-truth NVV label falls within the top-10 predictions with a probability of 92.2\%, with more than half ranking within the top-2. We carefully formulate validation rules that reasonably relax probability thresholds for rare label classes. Evaluations on real conversations alongside negatively sampled random insertions demonstrate that the model accepts 88.6\% of genuine dialogues while rejecting 91.3\% of unnatural insertions.
% while only around 7K correspond to NVV occurrences. 

\subsection{Speech Interaction Performance}

\subsubsection{Training Configuration and Indicators}

We conduct training on two speech interaction models, VocalNet and VITA-Qinyu, with the same Qwen3-8B backbone~\cite{wang2025vocalnet, xu2026vita}. We first perform S2T-SFT for 3 epochs, followed by GRPO training involving 3K samples. For speech response generation, we conduct 1 epoch of TTS training and 3 epochs of S2S training using the same data as in the S2T stage, with the corresponding speech responses synthesized by Minimax-HD 2.8. For evaluation, we primarily assess the appropriateness of generated NVV events, following the same criteria used for corpus filtering. We additionally consider consistency with the reference annotations as a complementary metric, while acknowledging that a given query may admit multiple natural responses with different NVV labels. For the vocalization quality, 5 human raters are instructed to evaluate in an NV-MOS (mean opinion score) manner on a scale of 1-5. UTMOS~\cite{saeki2022utmos} is reported as overall speech quality, and we transcribe the responses using Qwen3-ASR-1.7B~\cite{shi2026qwen3} and calculate the character error rate (CER) for linguistic speech, as the insertion of NVV events may affect the pronunciation of surrounding linguistic content.

\vspace{-2mm}
\subsubsection{NV Position Performance}

As shown in Table~\ref{tab: nvv_position}, after S2T-SFT, both models acquire the ability to generate NVV events that conform to conversational norms, with more than 85\% of complete responses containing only appropriate NVV events. Performance is even stronger in multi-turn role-play dialogue. NVV-aware GRPO yields consistent improvements, particularly for empathy-oriented interactions. For VocalNet, token-level NVV appropriateness increases from 92.5\% to 96.8\%, while both models achieve response-level appropriateness above 90\% on this dataset. Across the four sets of metrics corresponding to the two datasets, we observe consistent improvements, except for fluctuations in consistency with the ground-truth annotations, which mainly result from changes in response styles. S2S-SFT affects the two models differently. VocalNet preserves the capabilities of its backbone by fine-tuning only the speech generator, whereas VITA-Qinyu exposes additional LLM parameters due to its native architecture, resulting in a certain degree of performance degradation.

\begin{table}[t]
\centering
\resizebox{0.5\textwidth}{!}{
\begin{tabular}{lccccccc}
\toprule
 \multirow{2}{*}{\textbf{Model}} & \multicolumn{3}{c}{\textbf{Empathy Interaction}} & \multicolumn{3}{c}{\textbf{Role-play Dialogue}}  \\  \cmidrule(lr){2-4} \cmidrule(lr){5-7} 
 & NV-MOS & UTMOS & CER (\%) & NV-MOS & UTMOS & CER (\%) \\   
\midrule
VocalNet-NV & 3.78 & 2.85 & 1.20 & 3.94  & 3.04 & 1.78 \\
VITA-NV & 3.84 & 3.54 & 1.18 & 4.08 & 3.64 & 1.70 \\
\bottomrule
\end{tabular}
}
\caption{Speech quality evaluation of model responses.}
\label{tab:contextual_biasing}
\vspace{-4mm}
\end{table}

% Lixin 共情 vita3.68 vocalnet3.76  角色扮演vita3.66 vocalnet 3.93 
% Heyang 共情 VITA-NV 4.34 VocalNet 3.89 角色扮演  VITA-NV 4.48    VocalNet 4.16
% Jiayi 共情 VITA 4.45 VocalNet 4.50 角色扮演 VITA 4.43 VocalNet 4.27
% Zhen 共情 VITA 2.85 VocalNet 2.80 角色扮演 VITA 3.10 VocalNet 3.10
% Yuxin 共情  VITA 3.89 VocalNet 3.97 角色扮演 VITA 4.75 VocalNet 4.23
 
\subsubsection{NV Vocalization Quality}

VITA-NV consistently outperforms VocalNet-NV across all three evaluation metrics. It achieves an NV-MOS above 4 in role-play, indicating that it generates highly natural and fluent NVV events. VocalNet performs slightly worse, while a more noticeable inferiority is in the overall speech quality. VITA-NV achieves a UTMOS score above 3.5, while VocalNet sustains a score around 3.0. For linguistic pronunciation, both models maintain strong performance, with CER remaining below 2\%. In the calculation, we exclude text corresponding to discourse interjections to avoid the influence of NVV transcription on the evaluation metrics. These results suggest that, under the evaluated architectures, single-codebook modeling has limited capacity for representing newly introduced vocal expressions while preserving strong speech quality, whereas multi-codebook reconstruction provides a more effective approach for high-quality speech generation.

%% file: latex/5th_conclusion.tex
\vspace{-2mm}

\section{Conclusion}
\label{sec:conclusion}

Non-verbal vocalizations (NVVs) are an important component of natural and human-like speech interaction. We construct a high-quality data generation pipeline for an NVV-centric speech corpus, incorporate 13 major types of NVVs into Speech LLMs, and equip them with the ability to generate contextually appropriate NVV-aware responses through multi-stage training. In particular, NVV-aware GRPO further improves the appropriateness of NVV events in model outputs while mitigating the tendency to collapse into a limited subset of categories. Both subjective evaluations and objective metrics demonstrate the superior speech quality of VITA-NV, further highlighting the advantages of multi-codebook speech generation for natural and expressive speech interaction. These results suggest that modeling NVVs not only requires appropriate event selection and contextual integration, but also benefits from a speech generation paradigm with sufficient capacity to represent diverse vocal expressions.